\documentclass[]{fairmeta}

\usepackage{graphicx}
\graphicspath{ {./images/} }
\usepackage{amsmath}
\usepackage{amssymb}
\usepackage{booktabs}
\usepackage{algpseudocode}
\usepackage{mathtools}
\usepackage{listings}
\usepackage{xcolor}
\usepackage{siunitx}
\usepackage{multirow}
\usepackage{xspace}
\usepackage{enumitem}
\usepackage{bm}       
\usepackage{float}

\newcommand{\eg}{e.g.,\xspace}
\newcommand{\Our}{ICON\xspace}

\title{ICON: Incremental CONfidence for Joint Pose and Radiance Field Optimization}

\author[1]{Weiyao Wang}
\author[1,\dagger]{Pierre Gleize}
\author[1,\dagger]{Hao Tang}
\author[1]{Xingyu Chen}
\author[1]{Kevin J Liang}
\author[1]{Matt Feiszli}

\affiliation[1]{FAIR at Meta}

\contribution[\dagger]{Equal contribution.}

\abstract{
Neural Radiance Fields (NeRF) exhibit remarkable performance for Novel View Synthesis (NVS) given a set of 2D images. However, NeRF training requires accurate camera pose for each input view, typically obtained by Structure-from-Motion (SfM) pipelines. Recent works have attempted to relax this constraint, but they still often rely on decent initial poses which they can refine. Here we aim at removing the requirement for pose initialization. We present Incremental CONfidence (ICON), an optimization procedure for training NeRFs from 2D video frames. ICON only assumes smooth camera motion to estimate initial guess for poses.  Further, ICON introduces ``confidence": an adaptive measure of model quality used to dynamically reweight gradients. ICON relies on high-confidence poses to learn NeRF, and high-confidence 3D structure (as encoded by NeRF) to learn poses. We show that ICON, without prior pose initialization, achieves superior performance in both CO3D and HO3D versus methods which use SfM pose.
}

\date{\today}
\correspondence{Weiyao Wang at \email{weiyaowang@meta.com}; Matt Feiszli at \email{mdf@meta.com}}

\begin{document}

\maketitle

\section{Introduction}
\label{sec:intro}

\begin{figure}
\centering
{\small 
 \begin{subfigure}[c]{0.24\linewidth}
    {\includegraphics[width=\linewidth]{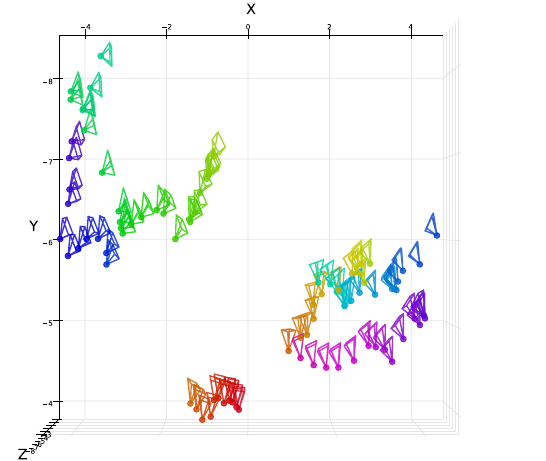}}
    \caption{BARF pose predictions}
 \end{subfigure}
  \begin{subfigure}[c]{0.24\linewidth}
    {\includegraphics[width=\linewidth]{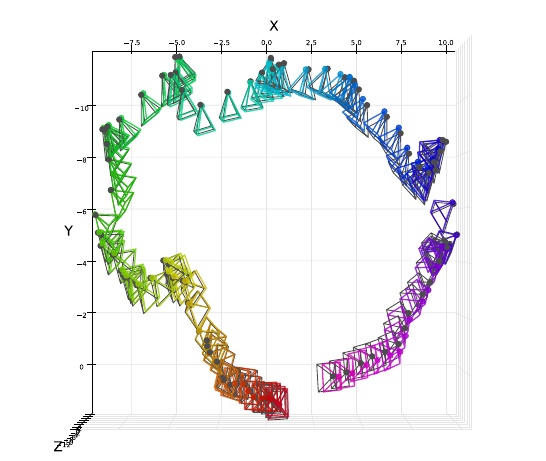}}
    \caption{\Our pose predictions}
 \end{subfigure}
 \begin{subfigure}[c]{0.24\linewidth}
    {\includegraphics[width=\linewidth]{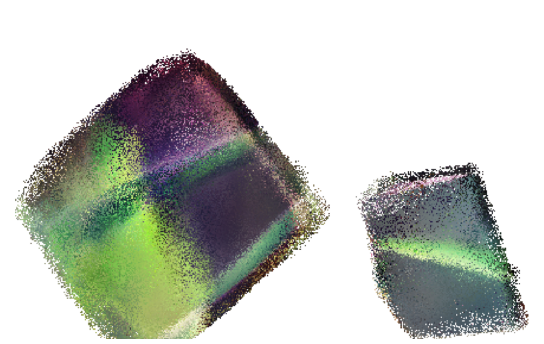}}
    \caption{BARF~\cite{lin2021barf} novel-view synthesis}
 \end{subfigure}
 \begin{subfigure}[c]{0.24\linewidth}
    {\includegraphics[width=\linewidth]{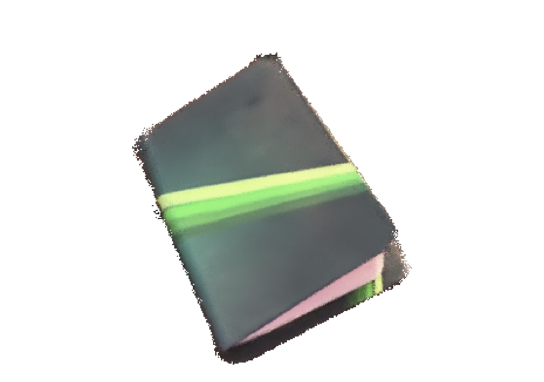}}
    \caption{\Our novel-view synthesis}
 \end{subfigure}

 \vspace{-8pt}
 \caption{{\bf Novel view and pose visualizations of \Our and BARF when no initial pose is available.} We train on a flyaround video of book from CO3D~\cite{reizenstein21co3d}. BARF trajectories exhibit fragmentation: camera poses split into two forward-facing clusters and create two books. \Our provides high-quality view synthesis and recovers poses very precisely. The colored triangle meshes represent \Our predicted poses and grey ones represent groundtruth. 
 }
 \label{fig:vis_teaser}
 }
 \vspace{-20pt}
\end{figure}

Robustly lifting objects into 3D from 2D videos is a challenging problem with wide-ranging applications. 
For example, advances in virtual, mixed, and augmented reality~\cite{pose4ar} are unlocking new interactions with virtual 3D objects; 3D object understanding is important for robotics as well (\eg manipulation~\cite{8263622,wen2021catgrasp,qi2023general} and learning-by-doing~\cite{Wen2022YouOD,cheng2023nodtamp}).


Bringing objects to 3D requires both extracting 3D structure and tracking 6DoF pose, but existing approaches have limitations.
Many~\cite{bundletrack2021,Azinovic_2022_CVPR,wen2023bundlesdf} rely on depth, which is a powerful signal for 3D reasoning.
However, accurate depth typically requires additional sensors (\eg stereo, LiDAR), which add cost, weight, and power consumption to a device, and is thus often not widely available.
Without this depth signal, these methods often fail.
Solving only half the problem is also common: 3D object reconstruction methods often assume pose~\cite{mildenhall2020nerf,reizenstein21co3d,Munkberg_2022_CVPR,Oechsle2021ICCV,sun2021neucon,wang2021neus,yariv2021volume}, and object pose estimation methods often assume a 3D model (\eg CAD)~\cite{pauwels2015simtrack, xiang2018posecnn, labbe2020cosypose}.
This chicken-and-egg problem often limits the applicability of these approaches. 

Here we aim to tackle both problems jointly, learning both an implicit 3D representation and per-frame camera poses from a single monocular RGB video. 
We supervise both 6DoF poses and reconstruction with a dense photometric loss, projecting the 3D representation onto the 2D input frames. Specifically, we represent objects/scenes as a Neural Radiance Field (NeRF) \cite{mildenhall2020nerf} to obtain 2D rendering.

While recent works \cite{yen2020inerf,lin2021barf,wang2021nerfmm,SCNeRF2021,lin2023icra:pnerf,sparf2023} have shown that poses can to some extent be 
(jointly) learned in this setting, they are most effective when used to refine initial poses with moderate noise. For example, \cite{wang2021nerfmm} shows they begin to fail when pose noise exceeds approximately 20 degrees of rotation error; more complex trajectories are unrecoverable. Indeed, these methods also fail on even moderately-complex trajectories, for example a full 360-degree flyaround of an object (Sec.~\ref{sec:experiments}). This means SfM preprocessing remains a prerequisite for constructing a radiance field.


One way forward would be to focus on the large-noise case, working to resolve larger pose changes. This is promising~\cite{meng2021gnerf}, but here we go the other way, and focus on the incremental case. This arises naturally in real-world settings where video is input, \eg embodied AI. We take inspiration from incremental SfM~\cite{schonberger2016structure} and SLAM~\cite{davison2003real}, training pose and NeRF jointly in an incremental setting. In this setup, the model takes a stream of video frames, one at a time. Leveraging a motion-smoothness prior, we initialize an incoming frame with the previous frame's pose. Information between frames is exchanged through view synthesis from NeRF. 

\begin{figure}
  \centering
    \includegraphics[width=0.65\linewidth]{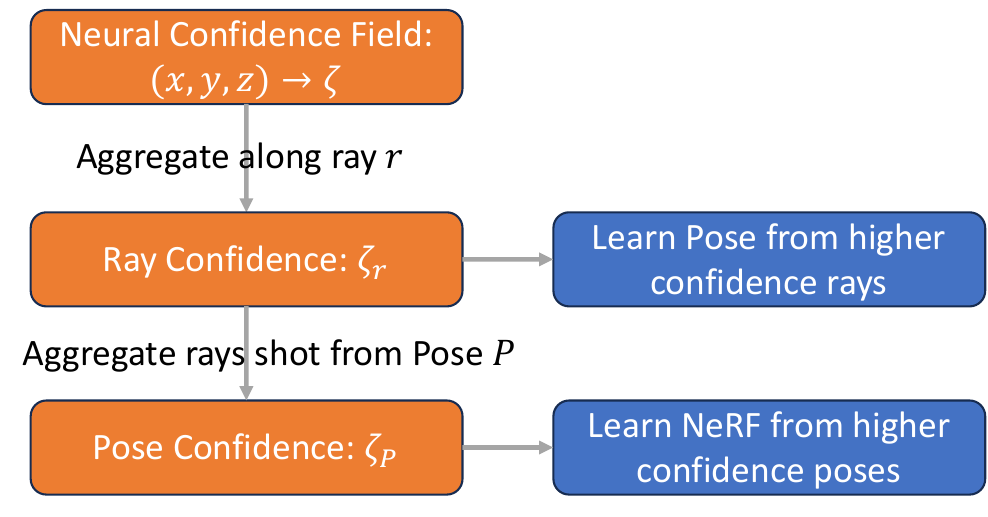}
  \vspace{-8pt}
  \caption{{\bf \Our overview}. \Our constructs a Neural Confidence field on top of NeRF to encode confidence $\zeta$ for each 3D location. The confidence is then used to guide the optimization process.}
  \label{fig:procedure}
  \vspace{-20pt}
\end{figure}

A major challenge comes from the interdependence between 3D structure and pose: high photometric error may be attributable to a poor 3D model despite good pose, or a large error in pose despite a good model.  We observe and analyze several interesting failure modes, including fragmentation, a generalization of the classical Bas-Relief ambiguity~\cite{belhumeur1999bas}, and overlapping registration (see Fig.~\ref{fig:failure_modes}). 

To address the difficulties, we propose \Our (Incremental CONfidence). The intuition is simple (Fig.~\ref{fig:procedure}): ``When pose is good, learn the NeRF; when the NeRF is good, learn pose."  ICON interpolates between these two regimes, using a measure of confidence obtained from photometric error, and maintaining a NeRF-style ``Neural Confidence Field" to store confidence in 3-space.  Confidence is also used as a signal to guide optimization; in particular it can help identify (and escape from) local minima. 

We perform quantitative evaluation of \Our on CO3D~\cite{reizenstein21co3d}, HO3D~\cite{hampali2020honnotate}, and LLFF~\cite{mildenhall2019local}.
While joint pose-and-3D baselines often fail catastrophically, \Our achieves strong performance on CO3D, comparable to NeRFs trained on COLMAP~\cite{schonberger2016structure} pose and surpassing a wide selection of baselines, such as DROID-SLAM~\cite{teed2021droid} and PoseDiffusion~\cite{posediff2023}. In addition, we evaluate on CO3D videos with background removed; this significantly increases the difficulty since background texture makes camera pose extraction easier. We note that this case (a single masked object in isolation) is quite valuable: success here means a method will work whether the camera is moving, the object is moving, or both. \Our achieves superior performance to NeRF+COLMAP pose and a wide selection of baselines
Finally, \Our outperforms RGB baselines and is comparable to SOTA RGB-D method BundleSDF~\cite{wen2023bundlesdf} on dynamic hand-held objects in HO3D.

To summarize, we make the following contributions:
\begin{enumerate}[noitemsep]
    \item We propose an incremental registration for joint pose and NeRF optimization. This setup removes the requirement for pose initialization in common video settings.
    \item We systematically study this incremental setup and discover several challenges. Based on the observations, we propose \Our, an optimization protocol based on confidence in spatial locations and poses. 
    \item We evaluate \Our with a focus on object-centric datasets. \Our is SOTA among RGB-only methods, and is even competitive with SOTA RGB-D methods.
\end{enumerate}
\section{Related Work}
\label{sec:related_work}

\textbf{Neural Radiance Field} (NeRF)~\cite{mildenhall2020nerf} is a powerful technique to represent 3D from posed 2D images for novel view synthesis. One major limitation of NeRF resides in its requirement for accurate camera poses. Recent works, including Nerf--~\cite{wang2021nerfmm}, BARF~\cite{lin2021barf}, SCNeRF~\cite{SCNeRF2021}, SiNeRF~\cite{Xia_2022_BMVC}, NeuROIC~\cite{10.1145/3528223.3530177}, IDR~\cite{yariv2020multiview}, GARF~\cite{chng2022gaussian} and SPARF~\cite{sparf2023} have attempted to relax this requirement by jointly optimizing poses and NeRF. Despite the promising direction, they work the best when refining noisy initial poses and are limited by the robustness of initial pose estimation methods. One direction the community takes to further reduce the dependency on pose is by adding additional components or signals for initial pose estimations, such as GANs~\cite{meng2021gnerf}, SLAM~\cite{rosinol2022nerf}, shape priors~\cite{zhang2021ners}, depth~\cite{bian2022nopenerf} and coarse annotations~\cite{boss2022-samurai}. We tackle this problem from a different angle, where we propose an incremental setup of joint NeRF and pose optimization. Our proposed method \Our does not use additional signals and achieve strong performance on challenging scenarios when camera poses are difficult to obtain. 

\noindent \textbf{Pose estimation (Object)} aims to infer the 6 Degrees-of-Freedom (DoF) pose of an object from image frames. The line of work can be classified into two main categories: image pose estimation \cite{xiang2018posecnn, labbe2020cosypose} and video pose tracking \cite{muller2021seeing, stoiber2022iterative, raftteed2020}, where the former mostly focuses on inferring pose from sparse frames and the latter takes the temporal information into consideration. However, many methods in video or image pose estimation assume known instance- or category-level object representations, including object CAD models \cite{xiang2018posecnn, labbe2020cosypose, labbe2022megapose, sundermeyer2018implicit, wang2019normalized, stoiber2022iterative, muller2021seeing} or pre-captured reference views with known poses \cite{liu2022gen6d, park2020latentfusion}. Recently, BundleTracks \cite{bundletrack2021} removes the need for such object priors, thus generalizing to pose tracking for unseen novel objects, and BundleSDF \cite{wen2023bundlesdf} improves pose tracking by constructing a neural representation for the object. However, both require depth information, limiting their applications.

\noindent\textbf{SLAM (Simultaneous Localization and Mapping)} builds a map of its environment while simultaneously determining its own location within that map \cite{mur2015orb,mur2017orb,davison2007monoslam, engel2014lsd, engel2017direct, klein2007parallel, zubizarreta2020direct}. While most SLAM methods focus on understanding camera pose movement in a static environment, object-centric SLAM \cite{mccormac2018fusion++, merrill2022symmetry, runz2018maskfusion, salas2013slam++, sharma2021compositional} focus on learning object pose in a dynamic environment. However, most of those methods require depth signal \cite{runz2018maskfusion, mccormac2018fusion++, merrill2022symmetry} and struggle with large occlusion or abrupt motion \cite{wen2023bundlesdf}.
\section{Method}
\label{sec:technical}

\Our takes streaming RGB video frames as input and produces 3D reconstructions and camera pose estimates. \Our incrementally registers each input frame to optimize 3D reconstruction guided by confidence: the 3D reconstruction is learned more from frames with high confidence pose, and pose relies on 3D-2D reprojection from higher confidence areas of the 3D reconstruction.

\subsection{Preliminaries: Neural Radiance Fields}

\Our relies on Neural Radiance Fields (NeRF) to represent a 3D reconstruction: NeRF encodes a 3D scene as a continuous 3D function through a multilayer perceptron (MLP) $f$ parameterized by $\Theta$: 3D point $x$ and viewing direction $d$ form the input $(\bm{x}, \bm{d})\in 
\bm{\mathbb{R}^5} \to (\textbf{c}, \sigma) \in \bm{\mathbb{R}^4}$, where $\textbf{c}\in \bm{\mathbb{R}^3}$ is the color and $\sigma$ is the opacity. To generate a 2D rendering of a scene at each pixel $p=(u,v)$ in image $\hat{I}_i$ from camera pose $P_i$, NeRF uses a rendering function $\mathcal{R}$ to aggregate the radiance along a ray shooting from the camera center $o_i$ position through the pixel $p$ into the volume:
\begin{equation}
\hat{I}_i(p) = \mathcal{R}(p,P_i|\Theta)=\int_{z_{\mathrm{near}}}^{z_{\mathrm{far}}} T(z) \sigma(\textbf{r}(z)) \textbf{c}(\textbf{r}(z), d) dz
\end{equation}
where $T(z)=\exp(-\int_{z_{\mathrm{near}}}^{z} \sigma(\textbf{r}(z)) dz)$ is the accumulated transmittance along the ray, and $\textbf{r}(z)=o_i+zd$ is the camera ray from origin $o_i$ through $p$, as determined by camera pose $P_i$. NeRF implements $\mathcal{R}$ by approximating the integral via sampled points along the ray, and is trained through a photometric loss between the groundtruth views $I_i$ and the rendered view $\hat{I}_i$ for all images $i=1,...,N$: 
\begin{equation}
  \Theta^*={\arg\min}_{\Theta} \mathcal{L}_{p}(\hat{I}|I,P),\text{where}~ \mathcal{L}_{p}(I, \hat{I})=\sum \|I_i-\hat{I}_i\|^2
\end{equation}


\begin{figure*}
  \centering
    \includegraphics[width=0.9\linewidth]{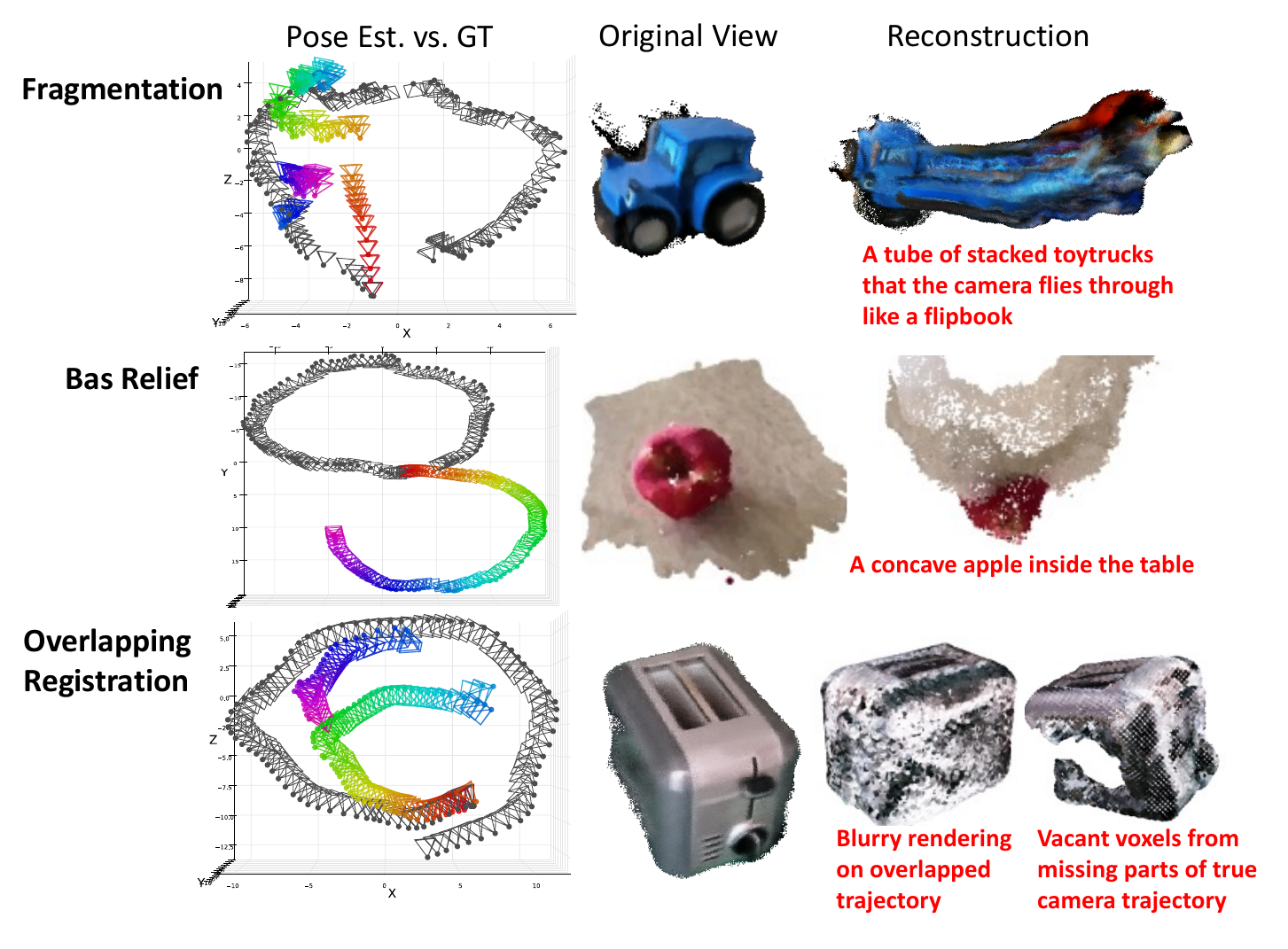}
  \vspace{-12pt}
  \caption{{\bf Three major failure modes of joint pose and NeRF optimization: fragmentation, Bas Relief, and overlapping registration}. The colored poses are predictions; grey poses are groundtruth. \textbf{Fragmentation}: Pose and NeRF break apart, producing separate, mutually invisible radiance fields. Here a tube of toytrucks is created, each occluding the next. Poses fly through this tube flipbook-style, each seeing a single toytruck. See also Fig.~\ref{fig:vis_teaser}, where completely independent reconstructions occur in different regions of 3-space.  \textbf{Bas Relief}: Due to an inherent ambiguity in RGB reconstruction, the model constructs a ``relief" by creating a concave apple inside the table, which results in camera trajectories inverted by 180 degrees. \textbf{Overlapping Registration}: Two subsets of the pose trajectory are trapped in a local minimum, incorrectly observing the same part of the radiance field
  , leading to blurry rendering and empty voxels. Here, one side of the toaster is blurry due to overlapping views, while the other has no views and is vacant.
  }
  \label{fig:failure_modes}
  \vspace{-12pt}
\end{figure*}

\subsection{Incremental frame registrations}

A major limitation for these joint pose and NeRF optimization methods is a requirement for good initial poses. If $\{P_i\}$ contain a diverse set of viewpoints and are initialized all from identity, these methods often collapse. For example, a simple but common collapsing solution is fragmentation: each frame creates its own fragmented 3D representation, all mutually invisible to the other views (\textbf{Fragmentation} fig.~\ref{fig:failure_modes}). Indeed, BARF~\cite{lin2021barf} collapses on all sequences of the CO3D dataset when the poses $\{P_i\}$ consist of a closed-loop flyaround (see Tab.~\ref{tab:co3d_full_scene}). As discussed in \cite{wang2021nerfmm}, when no pose prior is provided, a breaking point of 20 degree rotation difference for the whole trajectory is observed. 

To tackle this problem, we rely on a simple yet effective intuition: camera motions in videos are smooth. Therefore, given a frame $I_i$ in a video, its camera pose $P_i$ is likely to be close to $P_{i-1}$. We leverage this observation and propose to register frames incrementally following the temporal order. 

\noindent \textbf{Implementation.} At the start of training, we jointly optimize NeRF parameters $\Theta$ and poses $\{P_1, P_2\}$ from the first two frames $\{I_1, I_2\}$. After every $k$ iterations, we add a new frame $I_i$ and initialize its pose $P_i$ by $P_{i-1}$.  We freeze the learning rate on poses $\{P_i\}_{i=1}^N$ and NeRF $\Theta$ until all frames are registered. A learning rate decay schedule may be applied after all $N$ images are added.




\subsection{Confidence-Based Optimization}

The incremental registration process aims at providing good initialization for the camera poses. However, optimizing poses and NeRF using photometric losses is highly non-convex and contains many local minima~\cite{yen2020inerf,lin2023icra:pnerf}. 
In addition, an incorrectly optimized pose may provide misleading learning signals towards NeRF, increasing the possibility for poses to re-register incorrectly on already registered viewpoints (\textbf{Overlapping Registration} fig.~\ref{fig:failure_modes}). 


To tackle these, we propose a confidence-guided optimization schema. The intuition is simple: when a pose $P_i$ is confident, it should be trusted more to improve the learned NeRF $f(\Theta)$; when a ray sampled from $P_i$ contains locations that are confident, it should be weighted more to adjust the poses. When pose confidence drops dramatically for a new frame, it is likely that the pose got stuck in a local minima, so we perform a restart to re-register this pose. This is similar to the trial and error strategy of COLMAP~\cite{schonberger2016structure}. We next describe how we measure confidence for each pose $P_i$ and each point/viewing direction $(\bm{x},\bm{d})$ in 3D.

\noindent \textbf{Encoding confidence in 3D}.
We construct a Neural Confidence Field on top of NeRF: given an input 3D location and direction $(\bm{x},\bm{d})$, NeRF $f$ also predicts confidence $\zeta_{(\bm{x},\bm{d})}$. We add one fully-connected layer on top of the features, followed by a sigmoid, similar to the color prediction head. 

The confidence for a ray $\bm{r}$, is then aggregated through volumetric aggregation similar to opacity rendering: 
\begin{align}
    \zeta_{\bm{r}} &= (\int_{z_{\mathrm{near}}}^{z_{\mathrm{far}}} \mathcal{P}(z) dz) (\int_{z_{\mathrm{near}}}^{z_{\mathrm{far}}} \mathcal{P}(z) \zeta(\textbf{r}(z), d) dz) \nonumber \\
    &+ (1 - \int_{z_{\mathrm{near}}}^{z_{\mathrm{far}}} \mathcal{P}(z) dz)(\int_{z_{\mathrm{near}}}^{z_{\mathrm{far}}} \zeta(\textbf{r}(z), d) dz) 
\end{align}
where $\mathcal{P}(z) = T(z) \sigma(\textbf{r}(z))$. We note that the first term is more prominent when the pixel is opaque whereas the latter is more prominent for transparent pixels.



\noindent \textbf{Measuring confidence}. We measure confidence by how well a pixel reprojects in 2D through photometric error. Given a ray and its confidence $\zeta_{\bm{r}}$, we minimize $\mathcal{L}_{\mathrm{conf}}=\|e^{-\mathcal{E} / \tau}-\zeta_{\bm{r}}\|^2$, where $\mathcal{E}$ is the photometric error used to train NeRF and $\tau$ is a temperature parameter. $\mathcal{L}_{\mathrm{conf}}$ is only used to train the confidence head; gradient is stopped before NeRF parameters $\Theta$ or poses. 



\noindent \textbf{Pose confidence}. We compute pose confidence $\zeta_{P_i}$ for pose $P_i$ by aggregating confidence over rays sampled from $P_i$. At the start, $P_1$ has confidence 1 and others have confidence 0. During training, we use a momentum schedule to update pose confidence: at training iteration $t$, we sample $B$ rays $\{\bm{r}_j^i\}_{j=1}^B$ from pose $P_i$, and update confidence $\zeta_{P_i}^t$ as
\vspace{-3pt}
\begin{equation}
  \zeta_{P_i}^t = \beta \zeta_{P_i}^{t-1} + (1 - \beta) \frac{1}{B} \sum_{j=1}^B \zeta_{\bm{r}_j^i}  
\end{equation}
\vspace{-3pt}
The momentum $\beta$ is 0.9 in our experiments.

\textbf{Calibrating loss by confidence}. We use confidence to calibrate $\mathcal{L}$. Intuitively:
\begin{itemize}
    \item When we compute gradients for NeRF parameters $\Theta$, the loss is weighted by $\{\zeta_{P_i}\}$, the pose confidence.
    \item When we compute gradients for pose $\{P_i\}$, the per-ray loss is weighted by $\{\zeta_{\bm{r}}\}$, the ray confidence.  
\end{itemize}
\noindent At each step, we sample ray $\{\mathrm{r}_j^i\}_{j=1}^B$ from $P_i$. The loss is:
\vspace{-3pt}
\begin{gather}
\mathcal{L}_{\mathrm{NeRF}}(\Theta|\hat{P},I) = \sum_{i} (\sum_{j} \mathcal{L}(\bm{r}_j^i)) \zeta_{P_i}) / (\sum_{i,j} \zeta_{P_i}) \\
\mathcal{L}_{\mathrm{Pose}}(\hat{P}|\Theta,I) = \sum_{i, j} \mathcal{L}(\bm{r}_j^i) \zeta_{\bm{r}_j^i} / (\sum_{i,j} \zeta_{\bm{r}_j^i}) \\
\mathcal{L}_{\mathrm{all}}(\Theta, \hat{P}|I)=\mathcal{L}_{\mathrm{NeRF}}+\mathcal{L}_{\mathrm{Pose}}+\mathcal{L}_{\mathrm{conf}}
\end{gather}
\vspace{-3pt}

\noindent\textbf{Pose re-init}. Inspired by trial-and-error registration mechanisms in incremental SfM \cite{schonberger2016structure}, we do a re-initialization from the previous pose if a new image fails to register. We declare failure if we see an abrupt drop in confidence for a newly registered image: after we register $(I_i, P_i)$, 
we restart if new pose confidence $\zeta_{P_i}$ is less than $\lambda$ standard deviations of the mean of the $K$ previous pose confidences: $\zeta_{P_i} \le \mathrm{mean}(\{\zeta_{P_j}\}_{j=i-K}^{i-1}) - \lambda \cdot \mathrm{std}(\{\zeta_{P_j}\}_{j=i-K}^{i-1})$. We use $\lambda=2$ and $K=10$ throughout our experiments.

\subsection{Bas-Relief Ambiguity and Confidence-based Restart}

Bas-relief ambiguity~\cite{belhumeur1999bas}, and the related "hollow-face" optical illusion, are examples of fundamental ambiguity in recovering an object's 3D structure when objects that differ in shape produce identical images, perhaps under differing photometric conditions like lighting or shadow. For example, a surface with a round convex bump lit from the left may appear identical to the same surface with an concavity lit from the right. We refer generically to such situations as "Bas-Relief" solutions.  Human visual systems are known to employ strong priors (e.g. favoring convexity) to select a particular solution among multiple possibilities. 

We observe this phenomenon when jointly optimizing camera poses and NeRF, especially early in optimization when total camera motion is small. The model becomes stuck in a local minimum and cannot escape. For example, a concave version of the scene may be reconstructed when the groundtruth is a convex scene (see \textbf{Bas Relief} in Fig.~\ref{fig:failure_modes}). In this example, the camera movement is off by 180 degrees and moves in opposite directions compared to the groundtruth trajectory. We believe that simple priors, using cues like coarse depth, could help produce more human-like interpretations of natural scenes. However, for this study we avoid crafting priors, and remark that our confidence-based calibration of losses helps reduce this issue (16\% to 9\%).

We also observe that incorrect Bas Relief solutions generally have higher error and lower confidence; Relief solutions tend to be valid for a limited set of viewpoints and wider viewpoints become inconsistent. Hence we to propose a generic solution by adopting the restart strategy from incremental SfM. For example, COLMAP restarts to identify different initial pairs if the final reconstruction does not meet certain criteria (e.g. ratio of registered images). For us, we launch $K$ runs independently and measure the confidence after a fixed number of iterations. We pick the one with the highest confidence. In practice, we launch 3 runs and measure the confidence at 10\% of the training. 
\subsection{Confidence-based geometric constraint}

Following recent works~\cite{SCNeRF2021,sparf2023}, we add a geometric constraint to the optimization. Different from the ray-distance loss~\cite{SCNeRF2021} and depth consistency loss~\cite{sparf2023}, we adopt sampson distance~\cite{10.5555/861369}, similar to \cite{posediff2023}. We extract correspondence between a frame and its neighbors. We use SIFT~\cite{sift1999} features, primarily for fair comparison with COLMAP. At training time, for each pose $P_i$, we sample a pose $P_j$ in its neighbor, then compute Sampson distance:
\begin{equation}
\mathcal{L}_{\mathrm{Sampson}} = \frac{|x_i F x_j|}{|(x_i F)^1 + (x_i F)^2 + (F x_j)^1 + (F x_j)^2|}    
\end{equation}
where $F$ is the fundamental matrix between $P_i$ and $P_j$ and $(x_i F)^k$ indicates the $k$th element. 

\noindent \textbf{Loss calibration by confidence}. Although geometric cues help constrain the early optimization landscape, the correspondence pairs can be incorrect and/or not pixel-accurate, especially for objects with little texture. This causes the geometric constraint to be detrimental to \Our for obtaining precise poses and reconstructions. We rely on pose confidence $\zeta_{P_i}$ to weight the Sampson distance: for a pair of pose $P_i$ and $P_z$, weight by $1-\min(\zeta_{P_i},\zeta_{P_j})$. 





\section{Experiments}
\label{sec:experiments}


\noindent \textbf{Datasets}. We focus our study on Common Objects in 3D v2 (\textbf{CO3D}) dataset~\cite{reizenstein21co3d}, a large-scale dataset consisting of turn-table style videos of objects. Ground truth poses are obtained through COLMAP. We train on two versions of the dataset: \textbf{full-scene}, which uses the unmodified image frames (both object and background visible), and \textbf{object-only}, which removes the background leaving only foreground object pixels. We believe the object-only version is a more challenging yet meaningful evaluation set; in full-scene, objects are often placed on textured backgrounds where COLMAP can successfully extract poses. This implicitly equates object pose and camera pose, and this assumption breaks in dynamic scenes where both object and camera are moving. We use 18 categories specified by the dev set, with ``vase'' and ``donut'' removed due to symmetry (indistinguishable in the object-only setting). We select scenes with high COLMAP pose confidence for camera pose evaluation. We clean the masks using TrackAnything~\cite{yang2023track}; results on original masks are present in the supplementary. To demonstrate performance on dynamic objects, we additionally re-purpose \textbf{HO3D}~\cite{hampali2020honnotate} v2 to evaluate the camera pose tracking and view synthesis quality. HO3D consists of static camera RGBD videos capturing dynamic objects manipulated by human hands. We only use the RGB frames for \Our and select 8 clips (each around 200 frames) from 8 videos, each covering a different object. Finally, we show results on \textbf{LLFF}~\cite{mildenhall2019local}, a dataset with 8 forward-facing scenes commonly used for scene-level novel view synthesis, especially for NeRFs. 

\noindent \textbf{Architectures and Losses} Our architecture follows NeRF~\cite{mildenhall2020nerf} (no hierarchical sampling) and set the image's longer edge to 640. We use the standard MSE loss of NeRF. When using Sampson distance, it is weighted by $10^{-4}$. For the object-only settings in CO3D and HO3D, where object masks are available, we use MSE loss to supervise the opacity. For HO3D, we use hand masks when provided (7 out of 8 clips) to avoid sampling rays from occluded regions. 

\noindent \textbf{Training}. We use BARF~\cite{lin2021barf} settings and train for 200k iterations. For CO3D and HO3D, we skip every other frame to reduce training time, producing sequences around 100 frames. For \Our and its variants, we add a new frame every 1k iterations (CO3D/HO3D) / 500 iterations (LLFF) and freeze the learning rate (100k iterations for HO3D and CO3D, 30k for LLFF). Following BARF, we do not use positional encodings during registration and apply coarse-to-fine positional encoding after registration. 

\noindent \textbf{Evaluation}. Following \cite{lin2021barf}, we evaluate on the last part (typically 10\%) of each sequence. We measure camera pose quality with Absolute Trajectory Error (ATE)~\cite{zhang2018tutorial}, performing Umeyama alignment~\cite{umeyama1991least} of predicted camera centers with ground truth. ATE consists of a translation (ATE) and rotation (ATE$_{\mathrm{rot}}$) component, evaluating $l2$-distance between camera centers and angular distance between aligned cameras, respectively. For novel view synthesis, we run an additional test-time pose refinement, following standard practices in previous works~\cite{lin2021barf,wang2021nerfmm,yen2020inerf,sparf2023}. We use PSNR, LPIPS~\cite{8578166}, and SSIM as metrics.

\noindent \textbf{Baselines}. We build \Our on top of \textbf{BARF}~\cite{lin2021barf}, and compare against BARF for joint pose and NeRF optimization. For novel-view synthesis, we train NeRF with ground truth poses. For pose, we compare against a wide selection of baselines: \textbf{PoseDiff}~\cite{posediff2023} models SfM within a probabilistic pose diffusion framework; concurrent work \textbf{FlowCam}  FlowCAM~\cite{smith2023flowcam} solves pose from estimated 3D scene flow; \textbf{DROID}-SLAM~\cite{teed2021droid} is a SOTA end-to-end learning-based SLAM system. We also use their predicted poses to initialize and train NeRF. In addition, on object-only CO3D evaluation, we evaluate poses from state-of-the-art SfM pipeline \textbf{COLMAP}~\cite{schonberger2016structure} and an augment version of COLMAP~\cite{sarlin2019coarse} using learning-based features SuperPoint~\cite{DeTone2017SuperPointSI}+SuperGlue~\cite{sarlin2020superglue} (\textbf{COLMAP+SPSG}). Though \Our only uses \textit{RGB}, we include popular \textit{RGB-D} methods on HO3D, including DROID with ground truth depth input, \textbf{BundleTrack}~\cite{bundletrack2021} and state-of-the-art \textbf{BundleSDF}~\cite{wen2023bundlesdf}. 

\subsection{Full scene from CO3D}

\begin{figure*}
  \centering
    \includegraphics[width=0.9\linewidth]{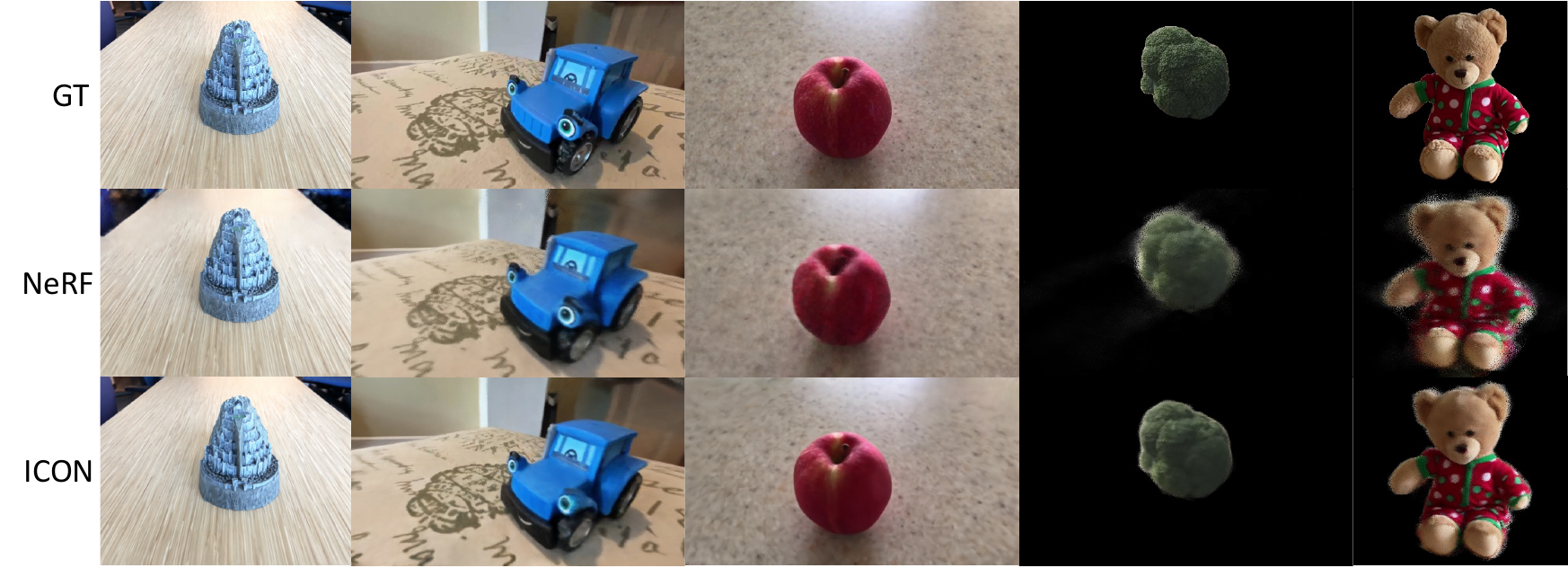}
  \vspace{-8pt}
  \caption{{\bf Novel view synthesis visualization of \Our without poses and NeRF trained with GT poses}. Despite having no pose priors, \Our renders novel views at comparable or higher quality. Results are taken from LLFF and CO3D. 
  }
  \label{fig:novel_view}
  \vspace{-10pt}
\end{figure*}

\begin{table}[t]
    \centering
    
    \begin{tabular}{cccccc}
\multicolumn{1}{l|}{}             & ATE            & ATE$_{rot}$   & PSNR           & SSIM           & LPIPS          \\ \hline
\multicolumn{6}{l}{Pose Source + NeRF}                                                                                \\
\multicolumn{1}{c|}{DROID}        & 0.431          & 8.92          & 17.19          & 0.526          & 0.541          \\
\multicolumn{1}{c|}{FLOW-CAM}     & 2.681          & 91.28         & 14.40          & 0.441          & 0.689          \\
\multicolumn{1}{c|}{PoseDiff}     & 1.973          & 27.25         & 18.82          & 0.563          & 0.520          \\ \hline
\multicolumn{1}{c|}{Groundtruth} & -              & -             & 21.03          & 0.575          & 0.629          \\ \hline
\multicolumn{6}{l}{Joint Pose + NeRF optimization}                                                                    \\
\multicolumn{1}{c|}{BARF}         & 6.215          & 114.63        & 12.77          & 0.401          & 0.871          \\
\multicolumn{1}{c|}{GT-Pose+BARF} & 0.417          & 3.77          & 19.33          & 0.558          & 0.647          \\ \hline
\multicolumn{1}{c|}{ICON (Ours)}  & \textbf{0.138} & \textbf{1.16} & \textbf{22.24} & \textbf{0.654} & \textbf{0.428}
\end{tabular}
\vspace{-8pt}
\caption{\textbf{Comparison on CO3D~\cite{reizenstein21co3d} full image scenes}. While baseline BARF may fail on CO3D due to larger camera motion overall, \Our can estimate poses very precisely and render novel views at quality similar or better than NeRF trained with GT poses.} \label{tab:co3d_full_scene}
\vspace{-12pt}
\end{table}

\noindent \textbf{\Our is strong on full-scene CO3D.} We compare ICON and baselines on full CO3D scenes in Table~\ref{tab:co3d_full_scene}. 
Without prior knowledge, BARF must initialize all camera poses as identity. CO3D's flyaround 
captures of objects result in camera pose variation that significantly exceeds the threshold after which BARF's performance collapses, with an ATE$_{\mathrm{rot}}$ exceeding 100 degrees. 
In contrast, ICON's incremental approach recovers significantly more precise camera poses (ATE of 0.137 and ATE$_{\mathrm{rot}}$ of 1.20), while also achieving better visual fidelity, both qualitatively and quantitatively, as measured by PSNR, SSIM, and LPIPS.
Interestingly, ICON still outperforms BARF \textit{even if BARF is provided with the ground truth poses at initialization}.  We originally proposed this setting as an upper bound, but we believe this result reflects instability in early iterations of BARF training: CO3D sequences are challenging compared to BARF benchmark scenes (e.g. synthetic dataset from \cite{mildenhall2020nerf}/forward facing LLFF). 
 Camera coverage is sparser, with more drastic lighting changes, and motion blur. Among the 18 scenes, BARF suffers from $\ge$ 10 degree ATE$_{\mathrm{rot}}$ in 4, dragging down the overall performance.

We also make several comparisons with NeRF~\cite{mildenhall2020nerf} and pose prediction methods. 
We provide NeRF with poses predicted by DROID-SLAM, FLOW-CAM, and PoseDiff, which rely on annotated poses to train or additional signals such as optical flow~\cite{raftteed2020}. 
However, our joint NeRF and pose training produces better pose estimates (as measured by ATE and ATE$_{\mathrm{rot}}$), and as a result, NeRF's novel view synthesis suffers in comparison. 
Even given CO3D's ground truth poses, ICON can outperform NeRF. 
While this may at first seem surprising, we point out that even the ``ground truth'' poses in CO3D are not true ground truth; they are generated with COLMAP, which is not perfect.
Additionally, in contrast to COLMAP, \Our's joint learning of NeRF and poses means that the estimated poses are specifically optimized to also maximize NeRF quality.
We hypothesize that this leads to poses more compatible for learning a NeRF, as reflected by the better performance we observe. Similar observations were presented in prior works~\cite{SCNeRF2021,meng2021gnerf}.

\subsection{Object-only on CO3D}
\begin{table}[t]
    \centering
    
    \begin{tabular}{cccccc}
\multicolumn{1}{l|}{}                & ATE            & ATE$_{rot}$   & PSNR           & SSIM           & LPIPS          \\ \hline
\multicolumn{6}{l}{Pose Source + NeRF}                                                                                   \\
\multicolumn{1}{c|}{DROID}           & 5.903          & 90.25         & 14.54          & 0.181          & 0.818          \\
\multicolumn{1}{c|}{FLOW-CAM}        & 6.700          & 120.52        & 13.08          & 0.127          & 0.886          \\
\multicolumn{1}{c|}{PoseDiff}        & 4.601          & 64.24         & 15.42          & 0.508          & 0.492          \\
\multicolumn{1}{c|}{Groundtruth}         & -              & -             & 20.77          & 0.718          & 0.301          \\ \hline
\multicolumn{6}{l}{COLMAP variants}                                                                                      \\
\multicolumn{1}{c|}{COLMAP(11)}      & 1.177          & 13.62         & \multicolumn{3}{c}{\multirow{3}{*}{-}}           \\
\multicolumn{1}{c|}{COLMAP-SPSG(11)} & 2.815          & 38.37         & \multicolumn{3}{c}{}                             \\
\multicolumn{1}{c|}{COLMAP-SPSG}     & 3.616          & 43.74         & \multicolumn{3}{c}{}                             \\ \hline
\multicolumn{6}{l}{Joint Pose + NeRF optimization}                                                                       \\
GT-Pose+BARF                         & 2.055          & 17.00         & 15.65          & 0.802          & 0.277          \\
BARF                                 & 6.522          & 114.97        & 8.22           & 0.772          & 0.370          \\ \hline
ICON (Ours)                                & \textbf{0.215} & \textbf{1.80} & \textbf{22.45} & \textbf{0.893} & \textbf{0.132}
\end{tabular}
\vspace{-8pt}
\caption{\textbf{Comparison on CO3D~\cite{reizenstein21co3d} object-only scenes without background}. Despite the challenges with background removal and failure from other methods, \Our can obtain poses at high precision and render novel views at high-quality. Since COLMAP only successfully registered more than 50\% of frames on 11 objects, we marked it with ``(11)" for comparison. The SPSG version of COLMAP registers for all scenes, and we include a datapoint on the 11 scenes subset that vanilla COLMAP succeeds.}
\vspace{-12pt}
\end{table}\label{tab:co3d_masked}

6DoF pose is inherently tricky to annotate, so past datasets often restrict motion to either the object or the camera; in the latter case, visually distinct backgrounds (\eg specially designed patterns, such as QR codes around the object) are often used to make pose trajectory reconstruction easier. 
These strategies however do not generalize to more in-the-wild video, especially when both an object and the background (or camera) are moving.
For this reason, we also perform evaluations on CO3D with the background masked out; in such a setting, algorithms are forced to only rely on object-based visual signal for estimating pose (Table~\ref{tab:co3d_masked}).

In this challenging setting, we again observe that BARF fails to estimate accurate poses, as the camera trajectory changes beyond what BARF can correct.
Additionally, the difficulty of this setting produces further deterioration of BARF's novel view synthesis.
However, we observe that \Our can still handle such videos, even without signal from the background.
This implies \Our is viable for joint pose estimation and 3D object reconstruction on more general videos, when the background cannot be relied on.

As with our full-scene CO3D experiments, we compare with methods for estimating pose, and how well those poses work when fed to a NeRF.
We observe that without being able to leverage the background, these methods struggle mightily. 
Pose prediction ATE and ATE$_{\mathrm{rot}}$ from DROID-SLAM in particular shoot up from 0.431 to 5.903 and 8.92 to 90.25, respectively. 
With poorer pose, the quality of the learned NeRFs are also correspondingly worse.

For pose in particular, we additionally evaluate COLMAP and its variant COLMAP-SPSG, which replaces SIFT~\cite{sift1999} with SuperPoint-SuperGlue~\cite{DeTone2017SuperPointSI,sarlin2020superglue}, on how they predict pose from just the foreground objects of CO3D.
We observe that COLMAP performs significantly worse when it cannot rely on background cues, far worse than ICON.
We believe this finding to be especially significant, as COLMAP is often considered the gold standard for camera pose alignment, and is often treated as ``ground truth" (as in CO3D).
This suggests our incrementally learned joint pose and NeRF optimization represents a promising new alternative for posing moving foreground objects, even if the background or camera is also moving.

\subsection{Hand-held dynamic objects on HO3D}

\begin{table}[t]
    \centering
    \begin{tabular}{cccccc}
\multicolumn{1}{l|}{}            & Input                                       & ATE   & ATE$_{\mathrm{rot}}$ & Trans & PSNR               \\ \hline
\multicolumn{1}{c|}{BARF}        & \multicolumn{1}{c|}{\multirow{2}{*}{RGB}}   & 0.135 & 122.38      & 0.580 & 5.72               \\
\multicolumn{1}{c|}{ICON}        & \multicolumn{1}{c|}{}                       & \underline{0.033} & \underline{8.07}        & \underline{0.049} & \textbf{16.24}              \\ \hline
\multicolumn{6}{l}{Baselines}                                                                                                \\ \hline
\multicolumn{1}{c|}{DROID}       & \multicolumn{1}{c|}{RGB}                    & 0.187 & 114.71      & 0.548 & \multirow{4}{*}{-} \\ \cline{2-2}
\multicolumn{1}{c|}{DROID}       & \multicolumn{1}{c|}{\multirow{3}{*}{RGB-D}} & 0.105 & 51.93       & 0.262 &                    \\
\multicolumn{1}{c|}{BundleTrack} & \multicolumn{1}{c|}{}                       & 0.046 & 29.45       & 0.158 &                    \\
\multicolumn{1}{c|}{BundleSDF}   & \multicolumn{1}{c|}{}                       & \textbf{0.021} & \textbf{6.82}        & \textbf{0.030} &                   
\end{tabular}
\vspace{-8pt}
\caption{\textbf{Comparison on HO3D}~\cite{hampali2020honnotate}. \Our works robustly against faster motion (vs CO3D), hand occlusion and lack of background information. In fact, despite only using RGB inputs, \Our can track poses at similar precision as SOTA RGB-D BundleSDF.}
    \label{tab:ho3d}
    \vspace{-12pt}
\end{table}

Understanding handheld objects is of particular importance to many applications, as the very nature of interaction often implies importance, and hands are often the source of object motion.
Pose and 3D reconstructions are key components of understanding objects, so the ability to generate them from videos of handheld interactions is of high utility.
We show results on HO3D~\cite{hampali2020honnotate} in Table~\ref{tab:ho3d}.

\begin{table*}[ht]
\centering
\resizebox{\linewidth}{!}{
\setlength\tabcolsep{1.5pt}
\begin{tabular}{clll|ccccc|ccccc|ccccc}
\multicolumn{1}{l}{} &                                &                                &                                 & \multicolumn{5}{c|}{CO3D-FullImg}                                                 & \multicolumn{5}{c|}{CO3D-No Background}                                           & \multicolumn{5}{c}{HO3D}                                                          \\
Incre                & \multicolumn{1}{c}{Geo.}       & \multicolumn{1}{c}{Calib.}     & \multicolumn{1}{c|}{Restart}    & ATE            & ATE$_{rot}$   & PSNR           & SSIM           & LPIPS          & ATE            & ATE$_{rot}$   & PSNR           & SSIM           & LPIPS          & ATE            & ATE$_{rot}$   & PSNR           & SSIM           & LPIPS          \\ \cline{5-19} 
\checkmark           & \multicolumn{1}{c}{\checkmark} & \multicolumn{1}{c}{\checkmark} & \multicolumn{1}{c|}{\checkmark} & \textbf{0.138} & \textbf{1.16} & \textbf{22.24} & \textbf{0.654} & \textbf{0.428} & \textbf{0.215} & \textbf{1.80} & \underline{22.45} & \textbf{0.893} & \textbf{0.132} & \textbf{0.033} & \textbf{8.07} & \textbf{16.24} & \textbf{0.863} & \textbf{0.164} \\
\checkmark           & \multicolumn{1}{c}{\checkmark} & \multicolumn{1}{c}{\checkmark} &                                 & 0.714          & 25.40         & 20.48          & 0.632          & 0.486          & \underline{0.224}          & \underline{1.86}          & \textbf{22.47}          & \underline{0.892}          & \textbf{0.132}          & 0.035          & 27.32         & 15.02          & 0.873          & 0.670          \\
\checkmark           &                                & \multicolumn{1}{c}{\checkmark} & \multicolumn{1}{c|}{\checkmark} & 1.691          & 28.95         & 18.66          & 0.565          & 0.556          & 0.340          & 3.91          & 21.92          & 0.887          & 0.140          & 0.032          & 19.19         & 14.51          & 0.866          & 0.184          \\
\checkmark           & \multicolumn{1}{c}{\checkmark} &                                &                                 & 1.283          & 36.82         & 19.05          & 0.567          & 0.562          & 0.972          & 15.94         & 21.03          & 0.875          & 0.163          & 0.046          & 30.50         & 12.86          & 0.863          & 0.290          \\
\checkmark           &                                &                                &                                 & 3.075          & 78.49         & 14.38          & 0.454          & 0.816          & 0.890          & 8.05          & 20.67          & 0.850          & 0.187          & 0.076          & 32.26         & 12.51          & 0.870          & 0.189          \\
\multicolumn{1}{l}{} &                                &                                &                                 & 6.215          & 114.63        & 12.77          & 0.401          & 0.871          & 6.522          & 114.97        & 8.22           & 0.772          & 0.370          & 0.307          & 131.16        & 7.45           & 0.82           & 0.29          
\end{tabular}
}
\vspace{-8pt}
\caption{\textbf{Ablation study by removing components when possible}. We remark that all designed component are critical for \Our. In addition, we didn't observe Bas Relief on the CO3D Object-Only (No Background) scenes, so the effect of Restart is minimal.}
\label{tab:ablation}
\vspace{-12pt}
\end{table*}

\begin{figure}
  \centering
    \includegraphics[width=0.65\linewidth]{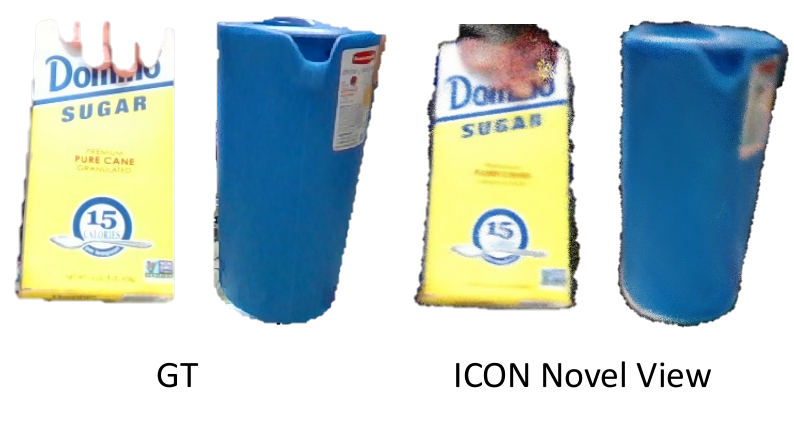}
  \vspace{-16pt}
  \caption{{\bf Visualization of \Our novel view synthesis on HO3D}. \Our can recover shapes and textures accurately.}
  \label{fig:ho3d_vis}
  \vspace{-10pt}
\end{figure}

Again, we primarily compare against BARF for joint object pose estimation and NeRF learning. Similar to CO3D object-only version, background is masked out since it moves differently than object. In addition, HO3D presents challenges with hand-occlusion and faster pose changes than CO3D. As with CO3D, we observe that BARF struggles to properly learn pose, especially with more drastic camera motion across nearby frames. 
On the other hand, \Our can perform well with these challenges: poses are predicted accurately (Tab~\ref{tab:ho3d}) and textures are rendered properly in novel views (Fig.~\ref{fig:ho3d_vis})

Several existing works~\cite{bundletrack2021, wen2023bundlesdf} addressing this problem additionally use depth, which provides a powerful signal for 3D object reconstruction and pose.
On the other hand, depth requires additional sensors and is not always available, and most visual media on the internet is RGB-only. 
Interestingly, we find that our results with \Our are competitive with state-of-the-art methods like BundleSDF which do require depth.  In addition, although we don't design or optimize \Our for mesh generation, we include a comparison on mesh by running an off-the-shelf MarchingCube~\cite{lorensen1987marching} algorithm. We follow the evaluation protocol in \cite{wen2023bundlesdf}, use ICP for alignment~\cite{besl1992method} and report Chamfer distnace. Despite not using depth signals, we found \Our provides competitive mesh quality (0.7cm) compared to BundleSDF (0.77cm).
We remark that BundleSDF's reconstruction performed poorly on one scene (2.39 cm); removing one worst scene for both method, BundleSDF and \Our achieved 0.54cm and 0.56cm. We believe that this represents the potential of monocular RGB-only methods for object pose estimation and 3D reconstruction.

\subsection{Ablation studies}

\textbf{What are the key components in \Our?}
We perform ablation studies to gain deeper insight why our proposed methodology leads to such significant improvements in Table~\ref{tab:ablation}, examining the impact of incremental frame registration (``Incre.''), as well as confidence-based geometric constraint (``Geo.''), loss calibration through confidence (``Calib.''), and restarts (``Restart'').
Note that the top row, with all options enabled, corresponds to our proposed ICON, while the bottom row (with none) is equivalent to BARF.
We find all the proposed techniques to be essential 

\noindent \textbf{\Our works on forward-facing scenes with minor camera motion.}
While much of our motivation and experiments center on the challenging setting of object-centric pose estimation and NeRF representations, we do not enforce any object-specific priors in our method. 
Our approach thus also generalizes to the scene images of LLFF~\cite{mildenhall2019local}, a common benchmark used by the wider NeRF community.
Compared to the type of videos in CO3D or HO3D, the images in LLFF tend to be forward-facing: the camera poses for each image have only mild differences.
Though easier, being able to recover camera poses in such settings is still important for wider applicability. 
We find that because the camera poses of LLFF only have limited variation, BARF initialized at identity is able to recover good poses and achieve good PSNR, SSIM, and LPIPS (Table~\ref{tab:LLFF}).
ICON, however, outperforms both BARF and a standard NeRF provided with ground truth poses.

\begin{table}[t]
    \centering
    \begin{tabular}{c|ccccc}
\multicolumn{1}{l|}{} & ATE   & ATE$_{\mathrm{rot}}$ & PSNR  & SSIM  & LPIPS \\ \hline
GT-Pose+NeRF          & -     & -           & 22.06 & 0.648 & 0.294 \\
BARF                  & 0.498 & 0.896       & 23.89 & 0.721 & 0.240 \\
ICON                  & \textbf{0.459} & \textbf{0.806}       & \textbf{24.23} & \textbf{0.731} & \textbf{0.221}
\end{tabular}
\vspace{-8pt}
\caption{\textbf{Comparison on LLFF~\cite{mildenhall2019local} dataset}. When camera poses have minor or mild motion, BARF works well with identity pose initialization and \Our performs slightly better. ATE is scaled by 100.}
    \label{tab:LLFF}
    \vspace{-12pt}
\end{table}


\section{Conclusion}
\label{sec:conclusion}

We proposed to study joint pose and NeRF optimization in an incremental setup and highlighted interesting and important challenges in this setting. To tackle them, we have designed \Our, a novel confidence-based optimization procedure. The strong empirical performance across multiple datasets suggests that \Our essentially removes the requirement for pose initialization in common videos. Although our focus is on object-centric scenarios, there are no priors or heuristics that rule out other settings.  \Our's LLFF and full-scene CO3D results are strong and show promise for more general types of video input, such as scene reconstruction from moving cameras (\eg egocentric~\cite{grauman2022ego4d}).

\clearpage
\newpage
\bibliographystyle{assets/plainnat}
\bibliography{main}

\clearpage
\newpage
\beginappendix

\section{Per-scene performance breakdown} \label{sec:breakdown}

We expand \Our results presented in main paper in section3 on CO3D full scene, CO3D object-only and HO3D~\cite{hampali2020honnotate} to document per-scene performance. Results are summarized in Tab.~\ref{tab:co3d_full_breakdown}, Tab.~\ref{tab:co3d_mask_breakdown} and Tab.~\ref{tab:ho3d_breakdown}.

\begin{table}[ht]
\centering
\small
\setlength\tabcolsep{1.5pt}
\begin{tabular}{llccccc}
Category & Scene & \multicolumn{1}{l}{ATE} & \multicolumn{1}{l}{ATE$_{rot}$} & \multicolumn{1}{l}{PSNR} & \multicolumn{1}{l}{SSIM} & \multicolumn{1}{l}{LPIPS} \\ \hline
apple      & 189\_20393\_38136  & 0.027 & 0.09 & 24.83 & 0.74 & 0.32 \\
ball       & 123\_14363\_28981  & 0.454 & 2.31 & 16.43 & 0.43 & 0.74 \\
bench      & 415\_57121\_110109 & 0.002 & 0.12 & 26.03 & 0.69 & 0.33 \\
book       & 247\_26469\_51778  & 0.219 & 1.41 & 26.79 & 0.76 & 0.30 \\
bowl       & 69\_5376\_12833    & 0.338 & 2.02 & 15.33 & 0.35 & 0.68 \\
broccoli   & 372\_41112\_81867  & 0.022 & 0.14 & 26.40 & 0.79 & 0.35 \\
cake       & 374\_42274\_84517  & 0.040 & 0.31 & 23.85 & 0.76 & 0.26 \\
hydrant    & 167\_18184\_34441  & 0.092 & 0.69 & 19.05 & 0.54 & 0.49 \\
mouse      & 377\_43416\_86289  & 0.240 & 1.33 & 22.33 & 0.71 & 0.36 \\
orange     & 374\_42196\_84367  & 0.200 & 3.86 & 24.71 & 0.80 & 0.35 \\
plant      & 247\_26441\_50907  & 0.190 & 1.95 & 16.30 & 0.43 & 0.59 \\
remote     & 350\_36761\_68623  & 0.043 & 0.28 & 27.08 & 0.66 & 0.42 \\
skateboard & 245\_26182\_52130  & 0.061 & 0.34 & 21.37 & 0.67 & 0.58 \\
suitcase   & 109\_12965\_23647  & 0.110 & 1.37 & 17.77 & 0.61 & 0.48 \\
teddybear  & 34\_1479\_4753     & 0.050 & 0.55 & 24.08 & 0.76 & 0.32 \\
toaster    & 372\_41229\_82130  & 0.240 & 2.57 & 20.11 & 0.53 & 0.50 \\
toytrain   & 240\_25394\_51994  & 0.170 & 1.92 & 19.08 & 0.66 & 0.49 \\
toytruck   & 190\_20494\_39385  & 0.010 & 0.17 & 27.39 & 0.87 & 0.15 \\ \hline
Avg        &                    & 0.138 & 1.16 & 22.24 & 0.65 & 0.43
\end{tabular}
\caption{Per-scene performance of \Our on CO3D full scene evaluation.}
\label{tab:co3d_full_breakdown}
\end{table}

\begin{table}[ht]
\centering
\small
\setlength\tabcolsep{1.5pt}
\begin{tabular}{llccccc}
Category & Scene & \multicolumn{1}{l}{ATE} & \multicolumn{1}{l}{ATE$_{rot}$} & \multicolumn{1}{l}{PSNR} & \multicolumn{1}{l}{SSIM} & \multicolumn{1}{l}{LPIPS} \\ \hline
apple      & 189\_20393\_38136  & 0.255 & 1.70 & 26.59 & 0.95 & 0.06 \\
ball       & 123\_14363\_28981  & 0.450 & 2.54 & 20.27 & 0.93 & 0.09 \\
bench      & 415\_57121\_110109 & 0.183 & 1.22 & 24.26 & 0.80 & 0.19 \\
book       & 247\_26469\_51778  & 0.174 & 1.36 & 24.24 & 0.89 & 0.13 \\
bowl       & 69\_5376\_12833    & 0.637 & 4.66 & 16.91 & 0.94 & 0.09 \\
broccoli   & 372\_41112\_81867  & 0.201 & 1.65 & 24.63 & 0.93 & 0.09 \\
cake       & 374\_42274\_84517  & 0.058 & 0.46 & 21.53 & 0.91 & 0.12 \\
hydrant    & 167\_18184\_34441  & 0.150 & 1.05 & 23.86 & 0.92 & 0.12 \\
mouse      & 377\_43416\_86289  & 0.420 & 7.09 & 15.93 & 0.80 & 0.31 \\
orange     & 374\_42196\_84367  & 0.387 & 3.84 & 29.34 & 0.98 & 0.02 \\
plant      & 247\_26441\_50907  & 0.075 & 0.62 & 18.28 & 0.75 & 0.27 \\
remote     & 350\_36761\_68623  & 0.109 & 0.71 & 25.38 & 0.94 & 0.09 \\
skateboard & 245\_26182\_52130  & 0.194 & 1.50 & 19.51 & 0.81 & 0.18 \\
suitcase   & 109\_12965\_23647  & 0.082 & 0.78 & 21.17 & 0.89 & 0.18 \\
teddybear  & 34\_1479\_4753     & 0.053 & 0.42 & 24.56 & 0.91 & 0.10 \\
toaster    & 372\_41229\_82130  & 0.225 & 1.01 & 20.79 & 0.94 & 0.10 \\
toytrain   & 240\_25394\_51994  & 0.159 & 1.19 & 20.35 & 0.83 & 0.18 \\
toytruck   & 190\_20494\_39385  & 0.066 & 0.68 & 26.46 & 0.95 & 0.05 \\ \hline
Avg        &                    & 0.215 & 1.80 & 22.45 & 0.89 & 0.13
\end{tabular}
\caption{Per-scene performance of \Our on CO3D object-only evaluation.}
\label{tab:co3d_mask_breakdown}
\end{table}

\begin{table}[ht]
\centering
\small
\begin{tabular}{lccccc}
       & ATE   & ATE$_{rot}$ & Trans & PSNR  & CD(cm) \\ \hline
SiS1   & 0.028 & 3.80   & 0.017 & 19.13 & 0.23   \\
MC1    & 0.019 & 5.90   & 0.049 & 14.24 & 0.41   \\
ABF13  & 0.064 & 10.67  & 0.094 & 11.79 & 1.72   \\
GPMF12 & 0.029 & 11.23  & 0.056 & 16.27 & 0.38   \\
ND2    & 0.027 & 7.18   & 0.015 & 20.06 & 0.50   \\
SM2    & 0.026 & 5.56   & 0.032 & 13.51 & 0.85   \\
SMu1   & 0.017 & 13.19  & 0.081 & 14.46 & 1.02   \\
AP13   & 0.058 & 7.06   & 0.046 & 20.42 & 0.50   \\ \hline
Avg    & 0.033 & 8.07   & 0.049 & 16.24 & 0.70  
\end{tabular}
\caption{Per-scene performance of \Our on HO3D evaluation. CD stands for Chamfer Distance, measuring mesh quality.}
\label{tab:ho3d_breakdown}
\end{table}

\section{Evaluating \Our on other CO3D categories} \label{sec:co3d_other}

In this section, we supplement the results reported in the main paper on CO3D~\cite{reizenstein21co3d}. We add a study using all the remaining 33 categories from CO3D and evaluate on the full scene. This makes it possible for us to include symmetric objects such as vase whose poses are indistinguishable in the object-only evaluation. Since no official subset is specified for these categories, we take top-4 instances from each category with highest camera pose confidence and randomly sample one instance for each category. It is worth noting that the ``ground-truth" camera poses are estimated by COLMAP, and may not be 100\% accurate, especially these categories are not part of the official benchmarking sets. We use the same (hyper-)parameters as the main paper benchmarking on the 18 categories.

\begin{table}[ht]
\centering
\small
\setlength\tabcolsep{1.5pt}
\begin{tabular}{llccccc}
Category      & Scene               & ATE   & ATE$_{rot}$ & PSNR  & SSIM & LPIPS \\ \hline
backpack      & 506\_72977\_141839  & 0.060 & 0.42   & 20.74 & 0.59 & 0.42  \\
banana        & 612\_97867\_196978  & 1.691 & 11.23  & 13.04 & 0.15 & 0.81  \\
baseballbat   & 375\_42661\_85494   & 0.791 & 7.83   & 13.92 & 0.61 & 0.68  \\
baseballglove & 350\_36909\_69272   & 0.054 & 0.72   & 20.52 & 0.43 & 0.62  \\
bicycle       & 62\_4324\_10701     & 0.700 & 5.94   & 15.22 & 0.19 & 0.69  \\
bottle        & 589\_88280\_175252  & 0.098 & 1.18   & 29.59 & 0.76 & 0.38  \\
car           & 439\_62880\_124254  & 0.765 & 4.43   & 11.40 & 0.32 & 0.87  \\
carrot        & 372\_40937\_81628   & 0.873 & 2.17   & 20.86 & 0.63 & 0.44  \\
cellphone     & 76\_7569\_15872     & 4.725 & 19.55  & 13.26 & 0.30 & 0.85  \\
chair         & 455\_64283\_126636  & 0.009 & 0.28   & 22.77 & 0.73 & 0.27  \\
couch         & 427\_59830\_115190  & 0.140 & 1.64   & 25.67 & 0.84 & 0.29  \\
cup           & 44\_2241\_6750      & 0.453 & 2.47   & 23.50 & 0.60 & 0.49  \\
donut         & 403\_52964\_103416  & 2.248 & 11.89  & 17.60 & 0.74 & 0.57  \\
frisbee       & 339\_35238\_64092   & 0.738 & 3.75   & 22.34 & 0.43 & 0.66  \\
hairdryer     & 378\_44249\_88180   & 0.022 & 0.16   & 25.84 & 0.82 & 0.33  \\
handbag       & 406\_54390\_105616  & 0.273 & 2.32   & 26.51 & 0.89 & 0.26  \\
hotdog        & 618\_100797\_202003 & 2.600 & 7.23   & 19.78 & 0.45 & 0.78  \\
keyboard      & 375\_42606\_85350   & 1.596 & 7.04   & 18.54 & 0.46 & 0.60  \\
kite          & 428\_60143\_116852  & 0.029 & 0.36   & 18.01 & 0.30 & 0.74  \\
laptop        & 378\_44295\_88252   & 1.128 & 7.92   & 15.04 & 0.36 & 0.59  \\
microwave     & 504\_72519\_140728  & 0.023 & 0.45   & 21.17 & 0.61 & 0.42  \\
motorcycle    & 367\_39692\_77422   & 0.006 & 0.14   & 26.52 & 0.78 & 0.30  \\
parkingmeter  & 483\_69196\_135585  & 0.136 & 2.48   & 17.24 & 0.56 & 0.56  \\
pizza         & 372\_41288\_82251   & 0.036 & 0.26   & 27.70 & 0.69 & 0.42  \\
sandwich      & 366\_39376\_76719   & 0.411 & 1.67   & 19.74 & 0.53 & 0.51  \\
stopsign      & 617\_99969\_199015  & 3.229 & 13.81  & 13.99 & 0.40 & 0.72  \\
toilet        & 605\_94579\_188112  & 0.252 & 5.48   & 18.53 & 0.69 & 0.41  \\
toybus        & 273\_29204\_56363   & 0.057 & 0.40   & 23.34 & 0.65 & 0.60  \\
toyplane      & 405\_53880\_105088  & 0.020 & 0.12   & 22.20 & 0.53 & 0.53  \\
tv            & 48\_2742\_8095      & 0.097 & 0.81   & 26.32 & 0.81 & 0.39  \\
umbrella      & 191\_20630\_39388   & 1.115 & 5.73   & 17.35 & 0.44 & 0.60  \\
vase          & 374\_41862\_83720   & 0.100 & 1.27   & 29.25 & 0.85 & 0.28  \\
wineglass     & 401\_51903\_101703  & 1.191 & 7.80   & 21.43 & 0.58 & 0.53  \\ \hline
Avg           &                     & 0.778 & 4.21   & 20.57 & 0.57 & 0.53 
\end{tabular}
\caption{Per-scene performance of \Our on other 33 categories in CO3D full-scene evaluation.}
\label{tab:co3d_other}
\end{table}

\begin{figure}
  \centering
    \includegraphics[width=0.65\linewidth]{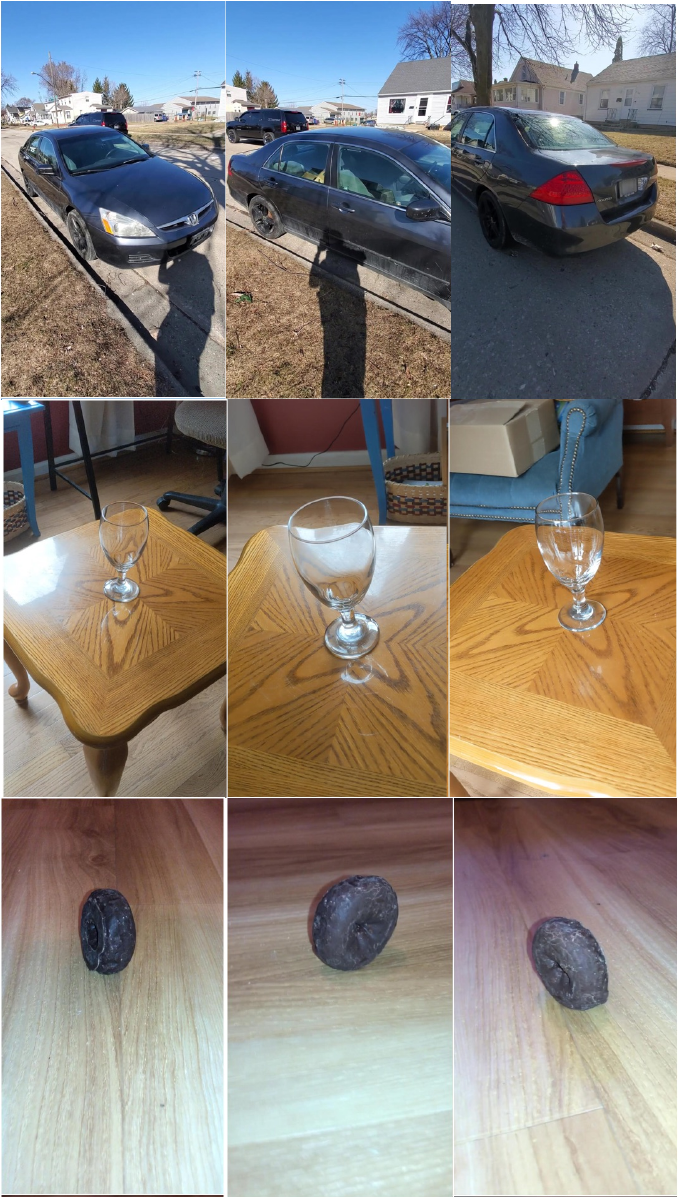}
  \caption{Scenes where \Our produces larger errors. \Our mainly suffer from scenes where photometric loss produces inconsistent supervisions. The \textbf{car} example consists of moving human shadow and reflective surface on the car. The \textbf{wineglass} example contains transparent surface and light reflections. The \textbf{donut} example contains inconsistent lighting, where the flash from the camera generates brighter color in the front and darken the back part. These inconsistencies in different viewpoints cause \Our to produce imprecise camera poses.}
  \label{fig:shadowed}
\end{figure}

We report the results in Tab~\ref{tab:co3d_other}. We observe that most objects achieve similar results as Tab~\ref{tab:co3d_full_breakdown}. However, there are a few objects where \Our yields imprecise poses, dragging down the average metrics. We believe there are two causes. First, \Our relies on photometric loss and may suffer from changes in the scenes. Many of the scenes where \Our has $\ge 3$ degree rotation error have moving shadows (either object or human), strong lighting change (from the builtin flash of the camera) or reflective surfaces. We show a few examples here in Fig.~\ref{fig:shadowed}. Second, the groundtruth poses used to evaluate the trajectory are generated by COLMAP, which may not be accurate, especially the categories not included in the official benchmarking sets.

\section{Evaluation on ScanNet} \label{sec:scannet}

\Our focuses our study on object-centric videos such as CO3D and HO3D. However, \Our does not apply specific design tailored towards object that prevents it to work on other types of videos. Here, we include a preliminary study by benchmarking \Our on ScanNet~\cite{dai2017scannet}. We randomly sample 10 out of 20 scenes in ScanNet test set and use a clip of 200 frames with a stride of 2. Scenes with NaN value in camera poses are removed when we sample scenes. 

We report camera pose quality following prior works~\cite{zhao2022particlesfm} using Relative Pose Error (RPE) on rotation and Absolute Trajectory Error (ATE (m)) for translation. We follow \cite{zhao2022particlesfm} to not use ATE$_{rot}$ because some trajectories in ScanNet has very small translation and aligning the trajectory then evaluate rotation may not be reliable. 

We do not change \textit{any} (hyper-)parameters used in CO3D full scene training for \Our to stress test the system on the significantly different scenarios in ScanNet. We include four methods designed to work well on ScanNet for comparison: TartanVO~\cite{tartanvo2020corl}, COLMAP~\cite{schonberger2016structure}, DROID-SLAM~\cite{teed2021droid} and current state-of-the-art method ParticleSfM~\cite{zhao2022particlesfm}. We note that COLMAP and ParticleSfM may fail to perform well when running only on the short clip, so we run them on the entire video and report the results on the clip. In addition, as noted in \cite{zhao2022particlesfm}, since COLMAP often fail on many ScanNet scenes, we use a tuned version following ~\cite{tschernezki21neuraldiff}.

\begin{table}[ht]
\centering
\small
\setlength\tabcolsep{2.5pt}
\begin{tabular}{l|ccccc}
            & \multicolumn{1}{l}{TartanVO} & \multicolumn{1}{l}{DROID} & \multicolumn{1}{l}{COLMAP} & \multicolumn{1}{l}{ParticleSfM} & \multicolumn{1}{l}{ICON} \\ \hline
RPE(degree) & 1.41                         & 0.56                      & 0.67                       & 0.34                            & 0.47                     \\
ATE(m)      & 0.198                        & 0.066                     & 0.091                      & 0.053                           & 0.092                   
\end{tabular}
\caption{Camera pose evaluation on ScanNet. Despite not optimized for ScanNet scenarios, \Our achieves competitive performance, ranking the second on RPE and third on ATE. The difference between \Our and state-of-the-art method is very small (0.13 degree on rotation and 0.039m on translation)}
\label{tab:scannet}
\end{table}

We report results in Tab~\ref{tab:scannet}. Despite having no tuning or change when transferring from CO3D, \Our achieves strong performance on ScanNet compared to the state-of-the-art methods designed to work well on ScanNet style videos. We believe this is a proof-of-concept that \Our can be generalized and adapted to other types of videos.

\section{Limitations and future directions} \label{sec:limit}

While \Our achieves strong performance to jointly optimize poses and NeRF, it has a few limitations. First, \Our strongly relies on photometric loss as supervision for both NeRF and poses. This relies on the assumption that the color is moderately consistent across different viewpoints. However, this assumption may break in real-world. Although \Our uses confidence to down-weight volumes with inconsistent photometric loss, it will produce imprecise poses (5 to 10 degree rotation error) due to the ambiguity. As shown in Tab~\ref{tab:co3d_other} and Fig~\ref{fig:shadowed}, \Our suffers from motion, reflective surfaces, transparency and strong lighting change. We believe leveraging features robust to these changes, such as DINO~\cite{caron2021emerging}, may help alleviate this issue.

In addition, \Our depends on gradient-based optimization through NeRF~\cite{mildenhall2020nerf}, which takes hours to train. We believe that combining \Our with more efficient modeling of 3-space will be a promising direction, such as PixelNeRF~\cite{yu2021pixelnerf} and FLOW-CAM~\cite{smith2023flowcam}. 


\end{document}